\documentclass[conference]{IEEEtran}
\IEEEoverridecommandlockouts
\usepackage{cite}
\usepackage{amsmath,amssymb,amsfonts}
\usepackage{algorithmic}
\usepackage{url}            
\usepackage{booktabs}       

\usepackage[linesnumbered,boxed,ruled,commentsnumbered]{algorithm2e}
\usepackage{graphicx}
\usepackage{textcomp}
\usepackage{booktabs}
\usepackage{xcolor}
\def\BibTeX{{\rm B\kern-.05em{\sc i\kern-.025em b}\kern-.08em
    T\kern-.1667em\lower.7ex\hbox{E}\kern-.125emX}}
\begin{document}

\title{Graded-Q Reinforcement Learning with Information-Enhanced State Encoder for Hierarchical Collaborative Multi-Vehicle Pursuit\\

\thanks{This work was supported by the National Natural Science Foundation of China (Grant No. 62071179) and project A02B01C01-201916D2}
}

\author{\IEEEauthorblockN{Yiying Yang,
Xinhang Li,
Zheng Yuan,
Qinwen Wang, 
Chen Xu, and
Lin Zhang
}
\IEEEauthorblockA{School of Artificial Intelligence, Beijing University of Posts and Telecommunications, Beijing, China}
\IEEEauthorblockA{\{yyying, lixinhang, yuanzheng, wangqinwen, chen.xu, zhanglin\}@bupt.edu.cn}
}


\maketitle

\begin{abstract}
The multi-vehicle pursuit (MVP), as a problem abstracted from various real-world scenarios, is becoming a hot research topic in Intelligent Transportation System (ITS). The combination of Artificial Intelligence (AI) and connected vehicles has greatly promoted the research development of MVP. However, existing works on MVP pay little attention to the importance of information exchange and cooperation among pursuing vehicles under the complex urban traffic environment. This paper proposed a graded-Q reinforcement learning with information-enhanced state encoder (GQRL-IESE) framework to address this hierarchical collaborative multi-vehicle pursuit (HCMVP) problem. In the GQRL-IESE, a cooperative graded Q scheme is proposed to facilitate the decision-making of pursuing vehicles to improve pursuing efficiency. Each pursuing vehicle further uses a deep Q network (DQN) to make decisions based on its encoded state. A coordinated Q optimizing network adjusts the individual decisions based on the current environment traffic information to obtain the global optimal action set. In addition, an information-enhanced state encoder is designed to extract critical information from multiple perspectives, and uses the attention mechanism to assist each pursuing vehicle in effectively determining the target. Extensive experimental results based on SUMO indicate that the total timestep of the proposed GQRL-IESE is less than other methods on average by 47.64$\%$, which demonstrates the excellent pursuing efficiency of the GQRL-IESE. Codes are outsourced in \emph{https://github.com/ANT-ITS/GQRL-IESE}.
\end{abstract}


\section{Introduction}
The Intelligent Transportation System (ITS), as an essential part of the smart city, is greatly facilitated by the development of emerging technologies. The Internet of Vehicles (IoVs) enables ITS to realize dynamic and intelligent management of traffic \cite{ITS1} \cite{ITS2}. Pursuit-evasion game (PEG), as a realistic problem for studying the self-learning and autonomous control of multiple agents, has been extensively studied in many fields, such as spacecraft control \cite{SC} and robot control \cite{RC}. Multi-vehicle pursuit (MVP), as an embodiment of PEG in ITS, has more conditional constraints, such as complex road structures, additional traffic participants, and traffic rules constraints. A patrol guide released by the New York City Police Department, representatively describes an MVP game, where multiple policy vehicles cooperate to capture single or multiple suspected vehicles \cite{a7}.

Regarding MVP, there have been some works on game theory-based methods. 
\cite{non1} focused on the multi-player pursuit game with malicious pursuers and constructed a nonzero-sum game framework to learn pursuers with different emotional intentions to complete the task. 
\cite{non2} developed a model predictive control method to address the problem of limited information on the pursuers, in which each pursuer only focused on its opponents' information. 
\cite{non3} adopted the graph-theoretic method to learn the interaction between the perception-limited agents and set the Minmax strategy to maintain the safe operation when the system failed to reach the Nash equilibrium.
However, it is difficult for these methods to construct a suitable objective function, and these methods pay little attention to the cooperation among pursuers in the dynamic traffic environment, which directly affects the effectiveness of the pursuit.
\begin{figure*}[t]
\centering
\centerline{\includegraphics[width=\textwidth,height=9.5cm]{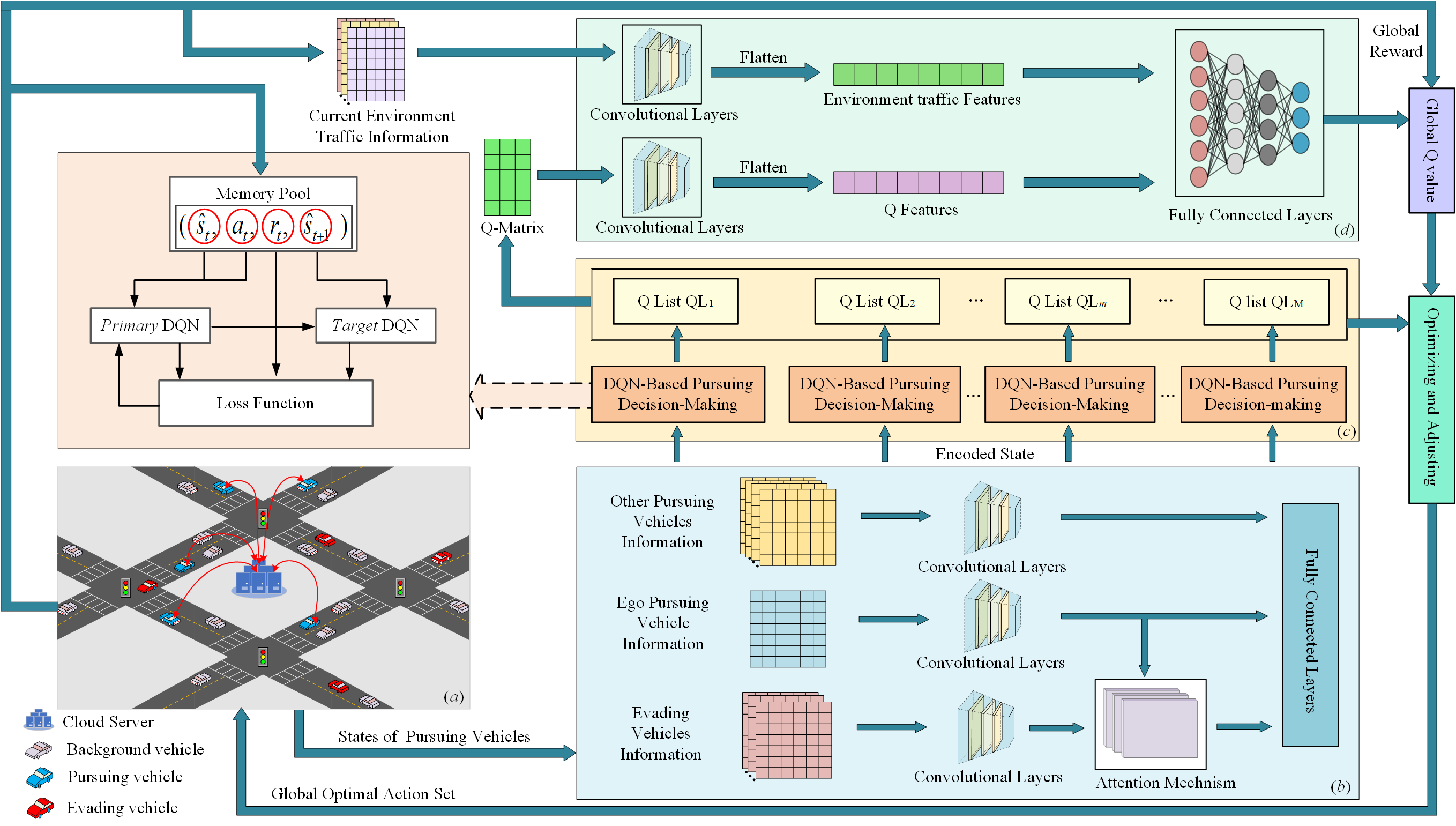}}
\caption{
The architecture of GQRL-IESE. (\textbf{a}) Complex urban traffic scene for HCMVP. (\textbf{b}) Information-enhanced state encoder (IESE). (\textbf{c}) DQN-based pursuing decision-making for multiple pursuing vehicles. (\textbf{d}) Coordinated Q optimizing network.
(\textbf{b}) encodes the state observed in (\textbf{a}) and inputs the encoded state into (\textbf{c}) for decision-making.
The decisions of the pursuing vehicles generated by (\textbf{c}) are not directly executed, while fed into (\textbf{d}) for evaluation, considering the current environment traffic information. The Q-matrix is optimized and adjusted according to the evaluation results of decision-making to obtain the current optimal action set.
}
\label{fig_1}
\end{figure*}

Cooperative multi-agent reinforcement learning (CoMARL) has been widely used in the coordinated control of multi-agent systems (MASs), such as traffic light control \cite{ATSC1},
and network resource allocation \cite{NSF}. CoMARL aims to maximize all agents' expected long-term common accumulative reward by learning a series of optimal policies or action sets \cite{co1}.
There is a growing research interest in applying CoMARL to MVP problem due to the powerful coordination mechanism and real-time decision-making ability of CoMARL. \cite{PE1} developed a probabilistic reward-based reinforcement learning (RL) method based on multi-agent deep deterministic policy gradient (MADDPG), where all pursuing agents are trained by a critic network, to accomplish the pursuit.
\cite{PE4} designed a target prediction network in the traditional general multi-agent reinforcement learning framework to more usefully assist the agents in decision-making.
\cite{PE5} introduced adversarial attack ticks and adversarial learning based on MADDPG to help agents learn more robust strategies.
\cite{PE2} added Transformer based on QMIX and learned historical observations from time and team, thereby promoting pursuers to learn cooperative pursuing strategies.
\cite{PE3} developed a CoMARL framework combining collaborative exploration and attention-QMIX to coordinately complete tasks, and the collaborative effectiveness of the CoMARL framework had been verified on a predator-prey scenario. 
However, these CoMARL methods on MVP are performed on the open or grid environment, and the complex traffic environments and traffic rules constraints will bring them new challenges.

In this paper, we propose a graded-Q reinforcement learning with information-enhanced state encoder framework (GQRL-IESE) for hierarchical collaborative multi-vehicle pursuit (HCMVP) under the complex urban environment. The architecture of the proposed GQRL-IESE is shown in Fig. \ref{fig_1}.
Compared with traditional RL, the proposed Graded-Q RL framework enhances the cooperative decision-making ability of agents in MASs.
In GQRL-IESE, an information-enhanced state encoder (IESE) is designed and implemented to encode complex states and extract effective information. Moreover, equipped with a cooperative graded-Q scheme, the GQRL-IESE coordinates the decisions of each pursuing vehicle to enable them to complete tasks cooperatively and efficiently. Furthermore, the main contributions of this paper are as follows:
\begin{itemize}
\item This paper proposes a graded-Q reinforcement learning with information-enhanced state encoder framework to address the HCMVP problem under the complex urban traffic environment.  
\item This paper designs an information-enhanced state encoder to extract crucial information from multi-dimension states of various pursuing participants, thus boosting the DQN-based decision-making of pursuing vehicles.
\item This paper proposes a cooperative graded-Q scheme to facilitate cooperation among pursuing vehicles, which introduces a coordinated Q optimizing network considering the current environment traffic information to promote the multi-agent pursuing policy.

\end{itemize}

The rest of this paper is organized as follows. Section \ref{II} presents an HCMVP problem and a detailed statement of the proposed GQRL-IESE for HCMVP. Section \ref{III} shows the structure of information-enhanced state encoder. The details of the proposed cooperative graded-Q scheme are given in Section \ref{IV}. Section \ref{V} conducts experiments to verify the performance of GQRL-IESE, and Section \ref{VI} concludes this paper.


\section{An Information-Enhanced Cooperative Reinforcement Learning Framework for HCMVP}\label{II}
\subsection{HCMVP Problem Statement Under Complex Urban Traffic Environment}
This paper focuses on the HCMVP problem under the complex urban traffic environment.
Different from the traditional MVP problem, where the pursuing vehicle only makes pursuing decisions according to its own information, the HCMVP problem focuses on the hierarchical optimization of cooperation and decision-making among pursuing vehicles. In the HCMVP problem, each pursuing vehicle can obtain global position information of other pursuing vehicles and evading vehicles through vehicle-to-vehicle (V2V) or vehicle-to-infrastructure (V2I). The goal of the HCMVP problem is to coordinately control the pursuing vehicles to capture all evading vehicles with the minimum pursuing time. This necessitates a feasible and effective hierarchical collaborative scheme to address the HCMVP problem.

\begin{figure}[h]
\centering
\centerline{\includegraphics[width=0.5 \textwidth, height=0.2\textwidth]{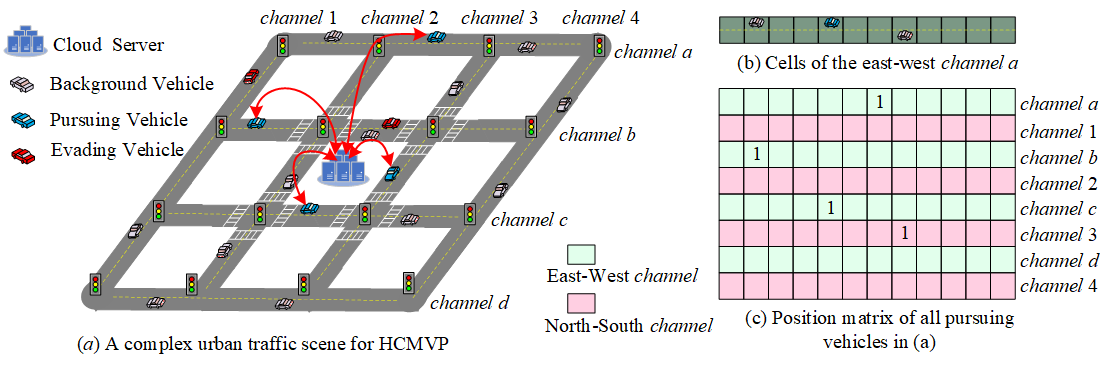}}
\caption{Global position information mapping of vehicles for HCMVP problem. (\textbf{a}) A complex urban traffic scene for HCMVP. (\textbf{b}) A process of dividing an east-west \emph{channel} into $K$ cells. (\textbf{c}) A position matrix of all pursuing vehicles in (\textbf{a}). 
}
\label{fig_2}
\end{figure}

We use Simulation of Urban Mobility \cite{SUMO} (SUMO) to simulate the complex urban traffic scene for HCMVP, as shown in Fig. \ref{fig_2}(a), which takes into account the constraints of urban traffic rules, additional traffic participants, and dynamic traffic flow.
In a closed traffic scene with $W \times W$ intersections, there are $M$ pursuing vehicles, $N$ evading vehicles ($M>N$), $B$ background vehicles, $L$ roads and a cloud server. Each road has bidirectional two lanes, and the traffic signal lights at each intersection are set as a fixed phase. To simulate the complex traffic situation and ensure collision-free driving, all vehicles in the scene are required to follow the traffic rules. The initial speed of the pursuing vehicles and evading vehicles are both $0$, and the values of their acceleration and maximum speed are fixed. 
\subsection{HCMVP Problem Modeling}
In the HCMVP problem, the pursuing vehicles aim to explore the optimal policy to maximize the accumulated reward. The essence of such a pursuing decision-making is well described by a markov decision process (MDP), defined by a tuple $\{ S,A,P,R,\gamma \}$. $S$ and $A$ are the finite set of the environment states and available actions, respectively. $R$ is a reward function, P is the state transportation probability function, and $\gamma$ is a discount factor to calculate accumulated rewards. At each time t, the agent observes its state $s_{t} \in S$, and selects an action $a_{t} \in A$. And then the agent obtains its next state $s_{t+1} \sim P\left(a_{t}, s_{t}\right)$, and the reward $r_{t} \sim R\left(a_{t}, s_{t}\right)$ from the environment by executing the action $a_{t}$.  

In the HCMVP problem, each pursuing vehicle strives to obtain the optimal policy to capture the target as quickly as possible. At each timestep, the pursuing vehicle needs to decide its actions. Unlike general MDPs, the pursuing vehicle in the HCMVP problem only executes the decisions when it reaches the intersection, due to the constraints of the traffic environment.
When the \emph{Euclidean distance} between any pursuing vehicle and the evading vehicle is less than the capture distance $d_{capture}$, the evading vehicle is captured, and then it will disappear from the scene. When all evading vehicles in the scene are captured, this pursuit is \emph{Done}. 
\subsection{Information-Enhanced Cooperative Reinforcement Learning Framework}
This paper proposes an information-enhanced cooperative reinforcement learning framework to address the HCMVP problem under the complex urban traffic environment. In the framework, we propose an information-enhanced state encoder (IESE), which extracts the critical information of the environment state from three perspectives, ego pursuing vehicle, other pursuing vehicles and the evading vehicles, to obtain the encoded state information.
Meanwhile, IESE adopts an attention mechanism, which facilitates the pursuing decision-making, and enables the pursuing vehicle to quickly and accurately determine its pursuing target with the minimum pursuing time, when it receives position information of multiple evading vehicles, thus improving the pursuing efficiency.

In the HCMVP problem, We propose a cooperative graded-Q scheme to promote cooperation among the pursuing vehicles. In the scheme, each pursuing vehicle generates decisions according to the received encoded state. 
The cloud server can collect the global information of the scene, including the positions of all pursuing and evading vehicles, and the global background traffic flow information.
Note that the pursuing vehicles send the decisions to the cloud server for optimization, instead of executing these decisions immediately. We design a coordinated Q optimizing network deployed on the cloud server. The coordinated Q optimizing network adjusts the received pursuing decisions according to the current environment traffic information, and generates the global optimal action set. The actions in the global optimal action set are separately distributed to each pursuing vehicle for executing, thus ensuring the global optimal decision-making of the pursuing vehicles.

This framework provides a feasible and effective solution for HCMVP in complex scenes, and facilitates collaboration among pursuing vehicles. In addition, the designed IESE helps the pursuing vehicles to extract the vital feature information of various pursuing participants, thereby improving the decision-making efficiency.

\section{Information-Enhanced State Encoder} \label{III}
This section proposes an information-enhanced state encoder to extract key features from various pursuing participants to facilitate pursuing decisions.
Section \ref{III-A} describes the modeling process of the global multi-dimension position information, and Section \ref{III-B} presents the specific structure of the proposed IESE.

\subsection{State Modeling} \label{III-A}
This paper defines a mapping matrix to effectively represent the road topology and the global position information of the vehicles.
The roads in the constructed complex traffic scene are divided into north-south roads and east-west roads. In a scene with $W \times W$ intersections, $W-1$ roads in the same direction that can be directly connected in space are defined as a \emph{channel}.
In order to represent the global position information of vehicles on the continuous roads, a \emph{channel} is directly divided into $K$ cells, as shown in Fig. \ref{fig_2}(b). Then a set of north-south channels $C_{sn}=\left\{{vc}_{1}, {vc}_{2}, \ldots, {vc}_{W}\right\}$, and a set of east-west channels $C_{ew}=\left\{{hc}_{1}, {hc}_{2}, \ldots, {hc}_{W}\right\}$ are obtained, where ${vc}_{i}$ and ${hc}_{i}$ represent a north-south \emph{channel} and an east-west \emph{channel} with $K$ cells, respectively. In order to effectively represent the topology of the road network, the elements in the two channels sets are cross-combined to obtain a ${2W} \times K$ mapping matrix ${MP}=\left[{hc}_{1}, {vc}_{1}, {hc}_{2}, {vc}_{2}, \ldots, {hc}_{W}, {vc}_{W}\right]$ of the road network. Fig. \ref{fig_2}(c) shows a position matrix of all pursuing vehicles in the complex urban traffic scene for HCMVP, which takes the representation of $MP$.

In order to assist the pursuing vehicles in cooperatively completing the pursuit task, the state of each pursuing vehicle should involve the global position information of other pursuing vehicles.
At timestep $t$, the state of the pursuing vehicle $m$ is defined as $s_{t}^{m}=\left\{{SF}_{t}^{m}, {SP}_{t}^{-m}, {SE}_{t}^{m}\right\}$, in which, ${SF}_{t}^{m}$, ${SP}_{t}^{-m}$ and ${SE}_{t}^{m}$ are the position matrix of the ego pursuing vehicle $m$, other pursuing vehicles and the evading vehicles, respectively. ${SF}_{t}^{m}$, ${SP}_{t}^{-m}$ and ${SE}_{t}^{m}$ take the representation of the mapping matrix $MP$. ${f}_{t, i, j}^{m}$ is the element of the $i^{th}$ row and $j^{th}$ column of the matrix ${SF}_{t}^{m}$, which indicates whether the pursuing vehicle $m$ is located in this cell. $p_{t, i, j}^{m}$ and $e_{t, i, j}^{m}$ are the elements of the $i^{th}$ row and $j^{th}$ column of the matrices ${SP}_{t}^{-m}$ and ${SE}_{t}^{m}$, respectively, which represent the number of other pursuing vehicles and evading vehicles in the current cell.

\subsection{Attention-Based State Encoder} \label{III-B}

\begin{figure}[h]
\centering
\centerline{\includegraphics[width=0.45 \textwidth, height=0.65 \textwidth]{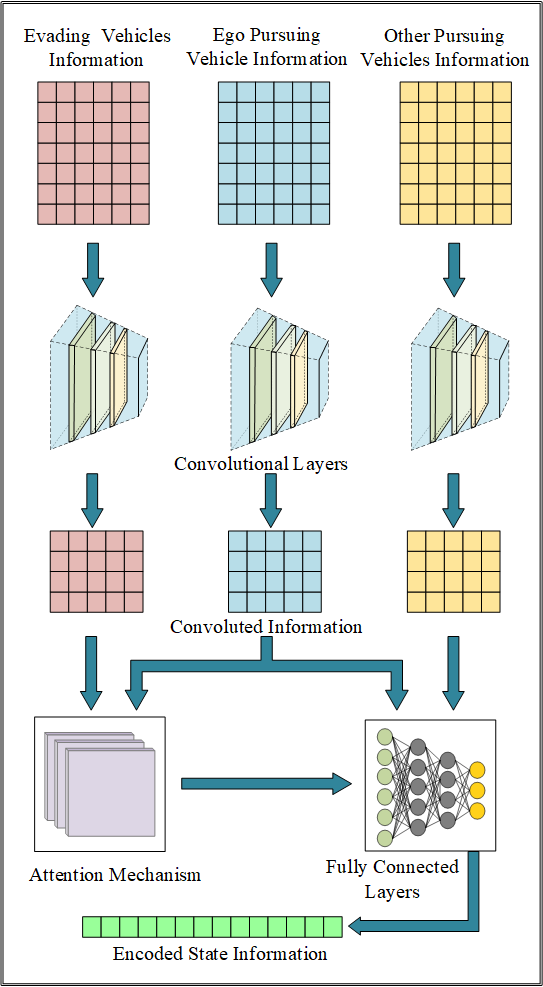}}
\caption{The structure of IESE.
IESE, consisting of convolutional layers, attention mechanism and fully connected layers, integrates the global position information of ego pursuing vehicle, other pursuing vehicles and evading vehicles.
}
\label{fig_3}
\end{figure}

In the HCMVP problem, the pursuing vehicle receives information from the cloud server and other pursuing vehicles in real time, which contains ineffective and interfering information, thus causing a great negative impact on the decision-making of the pursuing vehicle.
In this paper, the information-enhanced state encoder (IESE) adopts the attention mechanism to facilitate the pursuing vehicle to determine its pursuing target. And IESE is designed to extract the crucial information of the state from three perspectives of ego pursuing vehicle, other pursuing vehicles, and evading vehicles. The structure of IESE is shown in Fig. \ref{fig_3}.
IESE provides more appropriate state characteristic information for each pursuing vehicle, thereby promoting pursuing efficiency and improving the system's stability.

The attention mechanism is used to help a pursuing vehicle quickly and accurately determine its pursuing target. In the attention mechanism, \emph{Source} consists of a series of $(\text {\emph{Key}, \emph{Value}})$ pairs. The purpose of the attention mechanism is to obtain the weight coefficient of each \emph{Key} corresponding to the \emph{Value}, by calculating the correlation between the \emph{Query} of the given target and each \emph{Key}, and then weighting the \emph{Value} to get the final attention. The calculation formula is
\begin{equation}
\begin{aligned}
&\text {Attention}((\text {Key, Value}), \text {Query})=\sum_{i=1}^{N} a_{i} * \text {Value}_{i}\\
&=\sum_{i=1}^{N} \frac{\exp \left(\operatorname{sim}\left(\text {Key}_{i}, \text {Query}\right)\right)}{\sum_{j} \exp \left(\operatorname{sim}\left(\text {Key}_{j}, \text {Query}\right)\right)} * \text {Value}_{i},
\end{aligned}
\end{equation}
in which, $N$ is the length of \emph{Source}, $a_i$ is the attention weight, and $\operatorname{sim}(\cdot)$ is the similarity calculation mechanism.

Each pursuing vehicle has an IESE for state encoding. First, the three position matrices in the state $s_t^m$ are convoluted to extract crucial feature information from various perspectives, respectively. Then, the convoluted position information of ego pursuing vehicle and all evading vehicles are input into the attention mechanism to
boost the pursuing vehicle to concentrate on the information of its target evading vehicle, thus acquiring the target preference information of ego pursuing vehicle. Finally, the convoluted position information of ego pursuing vehicle and other pursuing vehicles, and the target preference information of ego pursuing vehicle are input into the fully connected layers for feature integration to obtain the encoded state information $\hat{s}_{t}^{m}$.
IESE adequately extracts the vital state information from ego pursuing vehicle, other pursuing vehicles and the evading vehicles, helping the pursuing vehicle to quickly determine the target vehicle. In addition, the processing of other pursuing vehicles' position information effectively promotes cooperation among the pursuing vehicles.

\section{Cooperative Graded-Q Multi-Agent Reinforcement Learning} \label{IV}
This section proposes a cooperative graded-Q scheme to facilitate the cooperation among pursuing vehicles and boost the pursuing vehicles to complete the pursuit with the minimum pursuing time. Section \ref{IV-A} introduces the deep Q network deployed on the pursuing vehicle, which makes decisions according to the information it receives. Section \ref{IV-B} proposes the coordinated Q optimizing network to collaboratively optimize the Q list of each pursuing vehicle. Finally, the decision-making and training process of the proposed GQRL-IESE is detailed in Section \ref{IV-C}.

\subsection{DQN-Based Pursuing Decision-Making} \label{IV-A}

Deep Q network (DQN), as a value-based reinforcement learning method, has been widely used in discrete action decision-making.
In the cooperative graded-Q scheme, each pursuing vehicle uses DQN to make decisions according to the encoded state information it receives.

In order to improve the decision-making effectiveness of the pursuing vehicle, the encoded state is input to the DQN. DQN uses neural networks to fit the Q function, and outputs the Q value for each action, containing turning left, turning right and going straight at the next intersection.
The Q list of each pursuing vehicle is defined as ${QL}_{m} = \left(q_{\text {left }}, q_{\text {righ }}, q_{\text {stra }}\right)$, in which, $q_{\text {left}}$, $q_{\text {righ}}$ and $q_{\text {stra}}$ indicate the probability that the pursuing vehicle $m$ selects to turn left, turn right and go straight, respectively.

Each pursuing vehicle receives an individual reward $r_{t}^{m}$ at each timestep to motivate the pursuing vehicles to complete the pursuing task. The individual reward is mainly composed of three parts. 
1) Each pursuing vehicle receives a negative reward of $c_1$ at each timestep when it does not complete the pursuit.
2) When there are g pursuing vehicles capturing the same evading vehicle at the same time, each pursuing vehicle gets a positive reward of $\frac{c_2} {g}$.
3) To encourage pursuing vehicles to aggressively capture evading vehicles, each pursuing vehicle receives a reward ${r_{\text {dis}}}_{t}^{m}$, which is designed by evaluating the distance between the pursuing vehicle and the closest evading vehicle to it. The calculation formula of the reward is 
\begin{equation}
\begin{aligned}
{r_{\text {dis}}}_{t}^{m}=\beta \times {\min \left\{d_{t+1}^{m, n}-d_{t}^{m,n}\right\}},\\
m \in\{1,2, \ldots, M\} \text { and } n \in\{1,2, \ldots, N\},
\end{aligned}
\end{equation}
where $\beta$ is the reward factor, and $d_{t}^{m,n}$ is the \emph{Euclidean distance} between the pursuing vehicle $m$ and the evading vehicle $n$ at timestep $t$.

In order to accelerate the convergence of networks and ensure smooth updating, the DQN of each pursuing vehicle maintains two networks with exactly the same structure, \emph{primary} DQN ${Q}_{m}\left(\theta_{m}\right)$ with the parameters $\theta_{m}$ and \emph{target} DQN ${Q}_{m}\left(\theta_{m}^{\prime}\right)$ with the parameters $\theta_{m}^{\prime}$.
In DQN, the agent uses an $\epsilon$-greedy strategy to select random actions, where the agent selects a random action with $\epsilon$ probability, and selects an optimal action with $1-\epsilon$ probability based on the estimated Q value. 
$\theta_{m}$ is iteratively updated in every learning through stochastic gradient descent (SGD) using data randomly sampled from the experience replay buffer $D^m$, and
$\theta_{m}^{\prime}$ is reset using $\theta_{m}$ for per fixed learning. 
Let an experience sample $e_{t}^{m}$ in $D^m$ is $e_{t}^{m}=\left\langle \hat s_{t}^{m}, a_{t}^{m}, r_{t}^{m}, \hat s_{t+1}^{m}\right\rangle$, and the loss function of DQN is
\begin{equation}
\begin{aligned}
L\left(\theta_{m}\right)=E_{e_{t}^{m} \in D^{m}}\left[\left(r_{t}^{m}\right.\right.&+\gamma \max _{a} Q_{m}\left(\hat s_{t+1}^{m}, a ; \theta_{m}^{\prime}\right) \\
&\left.\left.-Q_{m}\left(\hat s_{t}^{m}, a_{t}^{m} ; \theta_{m}\right)\right)^{2}\right],
\label{lossDQN}
\end{aligned}
\end{equation}
in which, $\gamma$ is the discounted factor. 

\subsection{Coordinated Q Optimizing Network} \label{IV-B}
In the graded-Q scheme, it introduces a coordinated Q optimizing network to evaluate and collaboratively optimize the Q list output by each DQN, while considering the current environment traffic information.
The coordinated Q optimizing network is updated using supervised learning, and the global reward is set as the evaluation benchmark.
The global reward is the sum of the individual rewards of each agent.
\begin{equation}
R_{t}=\sum_{m=1}^{M} r_{t}^{m}.
\end{equation}

The coordinated Q optimizing network aims to obtain the global optimal action set for the pursuing vehicles to enhance the cooperation among pursuing vehicles, thus improving the pursuing efficiency.

The input of the coordinated Q optimizing network contains the current environment traffic information $x_t$ and the decision information generated by all agents currently ${QM}_t$.
The current environment traffic information is defined as $x_{t}=\left\{{SP}_{t}, {SE}_{t}, {BN}_{t}\right\}$, in which, ${SP}_{t}$ and ${SE}_{t}$ separately represent the position information of all pursuing and evading vehicles, which take the representation of $MP$, and ${BN}_{t}$ represents the current number of background vehicles in each lane.
The Q-matrix ${QM}_t$ integrates the individual Q lists generated by the DQNs, whose element in the $m^{th}$ row is denoted as ${QM}_t^m$, representing the individual Q list ${QL}_m$ of the $m^{th}$ agent.
The coordinated Q optimizing network outputs the optimized decision information of all agents ${QM}_{t}^{g}$.
We employ an exploration approach to find the optimal ${QM}_{t}^{g}$ that maximizes the global Q-value estimated by the optimizing network. Then we obtain the optimal action set $A_t^g$ according to ${QM}_{t}^{g}$.
And the obtained optimal action set is separately distributed to each pursuing vehicle to execute. The coordinated Q optimizing network is iteratively updated through SGD using experience randomly sampled from the experience memory pool $U$. An experience in $U$ is defined as $\left\langle x_{t}, {QM}_{t}, R_{t}\right\rangle$.
The optimizing network is committed to fitting the global reward, and the loss function of the network is
\begin{equation}
L\left(\theta^{g}\right)=E_{\left\langle x_{t}, {QM}_{t}, R_{t}\right\rangle \in U}\left[\left(R_{t}-{Q_{tot}}\left(x_{t}, {QM}_{t}; \theta^{g}\right)\right)^{2}\right],
\label{lossQ}
\end{equation}
in which, ${\theta}^g$ is the parameter of the coordinated Q optimizing network.

\subsection{Decision-Making and Training Process of GQRL-IESE}\label{IV-C}
In GQRL-IESE, each pursuing vehicle uses a DQN to make decisions based on its own encoded state. These decisions are sent to the coordinated Q optimizing network to be optimized and adjusted at the global level, instead of being directly executed. Then these optimized actions are separately issued to each pursuing vehicle for execution, thus enhancing the cooperation among the pursuing vehicles and improving the pursuing efficiency.
The following describes the decision-making and training process of GQRL-IESE, as shown in Algorithm \ref{Alg1}.

\begin{algorithm}[ht]\label{Alg1}
            \caption{The decision-making and training process of GQRL-IESE}
            \LinesNumbered
            Initialize the HCMVP environment\;
            Initialize the experience replay buffer set $\left\{ D^{m} \right\}$\;
            Initialize the experience memory pool $U$\;
            Initialize the parameters of DQN $\left\{{\theta}_{m}\right\}$\;
            Initialize the agents' state $\left\{ s_{1}^{m} \right\}$ and $x_{1}$\;
            {\For{$t = 1:T$}
            {
            \For{$m = 1:M$}
            {
            Feed $s_{t}^{m}$ to IESE and obtain $\hat s_{t}^{m}$\;
            Get ${QL}_{m}$ through DQN according to $\hat s_{t}^{m}$\;
            }
            Integrate $\left\{ {QL}_{m} \right\}$ to get ${QM}_{t}$\;
            Input ${x_{t}}$ and ${QM}_{t}$ to the coordinated Q optimizing network and obtain $Q_{tot}$\;
            \For{$m = 1:M$}
            {
            Adjust ${QM}_t^{m}$ to get ${QM}_t^{'}$\;
            Recalculate $Q_{tot}^{'}$\;
            \If{$Q_{tot}^{'}>Q_{tot}$}
            {
               $Q_{tot}^g=Q_{tot}^{'}$ \;
               ${QM}_t^g={QM}_t^{'}$
            }
            }
            Select $A_t^{g}$ according to ${QM}_t^{g}$\;
            Execute $A_t^{g}$, and then obtain $R_t$, $r_{t}$, $s_{t+1}$ and $x_{t+1}$\;
            Store $\left \langle{\hat{s}_{t}^{m}, a_{t}^{m}, r_{t}^{m}, \hat{s}_{t+1}^{m}}\right\rangle$ to $D^{m}$ and $\left\langle x_{t}, QM_{t}, R_{t}\right\rangle$ in $U$\;
            Randomly select samples from $D^{m}$ and update ${\theta}_{m}$ via \eqref{lossDQN}\;
            Randomly select experience from $U$ and update ${\theta}^{g}$ via \eqref{lossQ}\;
            \If{\emph{Done}}
            {Break\;}
            }
      }
\end{algorithm}

At timestep $t$, the coordinated Q optimizing network observes the current environment traffic information $x_t$, and the agent $m$ observes the state $s_t^m$ from the environment, which is fed to IESE to get its encoded state $\hat{s}_{t}^{m}$. Then the agent adopts DQN to obtain the individual Q list ${QL}_m$ based on $\hat{s}_{t}^{m}$. After all agents obtain their individual Q lists, these individual Q lists are integrated into the Q-matrix ${QM}_t$.
$x_t$ and ${QM}_t$ are input to the coordinated Q optimizing network to obtain the global Q value $Q_{tot}$ of the current individual actions.
In the coordinated Q optimizing network, we use an exploration approach to find the optimal action set $A_t^g$. Randomly adjust the action of an agent to obtain a new ${QM}_{t}^{'}$ and its corresponding $Q_{tot}^{'}$. We obtain the maximum $Q_{tot}^{g}$ and its corresponding optimal Q-matrix ${QM}_{t}^{g}$ through traversing. Then the optimal action set $A_t^g$ is obtained according to ${QM}_{t}^{g}$. The coordinated Q optimizing network distributes the actions in $A_t^g$ to each agent to execute. After executing $A_t^g$, the agents receive individual reward $r_t$ and the next timestep state $s_{t+1}$ from the environment, and the coordinated Q optimizing network receives the global reward $R_t$ and the next timestep environment traffic information $x_{t+1}$ from the environment.
Afterward $\left \langle{\hat{s}_{t}^{m}, a_{t}^{m}, r_{t}^{m}, \hat{s}_{t+1}^{m}}\right\rangle$ is stored in $D^m$ and $\left\langle x_{t}, QM_{t}, R_{t}\right\rangle$ is stored in $U$. Then samples are selected from $D^m$ to update ${\theta}_{m}$ according to \eqref{lossDQN}, and experience is selected from $U$ to update the coordinated Q optimizing network according to \eqref{lossQ}. 
Repeat the above learning process until the pursuit is \emph{Done}.

The proposed GQRL-IESE motivates the pursuing vehicle to execute the policy that maximizes the global reward, and facilitates collaborative decision-making among pursuing vehicles, thereby improving the global cooperation of the pursuing vehicles. Particularly, the IESE urges the pursuing vehicles to effectively extract key information from different pursuing participants, and boosts the pursuing vehicle to determine its pursuing target more quickly, thus improving the pursuing efficiency.

\section{Comparison and Analysis of Performance}\label{V}
\subsection{Simulation Settings}
This paper simulates a complex urban traffic scene for HCMVP based on SUMO. We construct a $3km \times 3km$ space with $4 \times 4$ intersections. In the scene, each evading vehicle randomly selects one of the preset routes as its own path, and all background vehicles randomly determine routes. Of course, all vehicles in the scene are required to obey traffic rules. The parameter settings are presented in Table \ref{table1}.
\begin{table}[htbp]
\caption{Parameter Settings}
\label{table1}
\scalebox{0.75}{
\centering
\begin{tabular}{cc|cc}
\toprule
Parameters                                          & Value         & Parameters                                            & Value \\ \hline
Number of pursuing vehicles $M$                       & 4             & Number of evading vehicles $N$                          & 2     \\ 
Number of background vehicles $B$                     & 50            & Maximum speed            & 20$\text{ }m/s$ \\
Maximum acceleration  & 0.8 $\text{ }m/s^2$           & Maximum deceleration   & 4.5$\text{ }m/s^2$   \\ 
Number of rods $L$ & 24  & Number of cells $K$                                   & 10             \\ 
Learning rate                                & 0.001         & $\gamma$                                              & 0.95 \\ 
$\epsilon$                                          & 0.01          & $\beta$                                               & 2     \\ 
$c_1$                                               & -0.01          & $c_2$                                                 & 10    \\  \bottomrule
\end{tabular}
}
\end{table}


\begin{table*}[]
\caption{Simulation Results}
\label{table2}
\begin{center}
\scalebox{1.25}{

\begin{tabular}{cccccccc}
\toprule
                & GQRL-IESE & QMIX     & DDPG    & MADDPG  & DQN      & IESE+DQN & GQRL    \\ \hline
Total Timestep & 175.00    & 455.00   & 264.00  & 216.00  & 402.00   & 477.00   & 294.00  \\
Total Reward    & 407.79    & -496.87  & 144.58  & 363.42  & -638.30  & -625.28  & 313.79  \\
Improvement with GQRL-IESE  & -    & 221.85\% & 64.54\% & 10.88\% & 256.53\% & 253.34\% & 23.05\% \\
Average Reward  & 2.33      & -1.09    & 0.55    & 1.68    & -1.59    & -1.31    & 1.07    \\
Improvement with GQRL-IESE    & -    & 146.86\% & 76.50\% & 27.80\% & 168.14\% & 156.26\% & 54.20\% \\ \bottomrule
\end{tabular}

}

\end{center}
\end{table*}

\subsection{Performance Comparison with Other Methods}
In this section, we compare the convergence and pursuing efficiency of the proposed GQRL-IESE with other methods, including QMIX, deep deterministic policy gradient (DDPG), MADDPG and DQN. 

\begin{figure}[h]
\centering
\centerline{\includegraphics[width=0.4\textwidth,height=4.7cm]{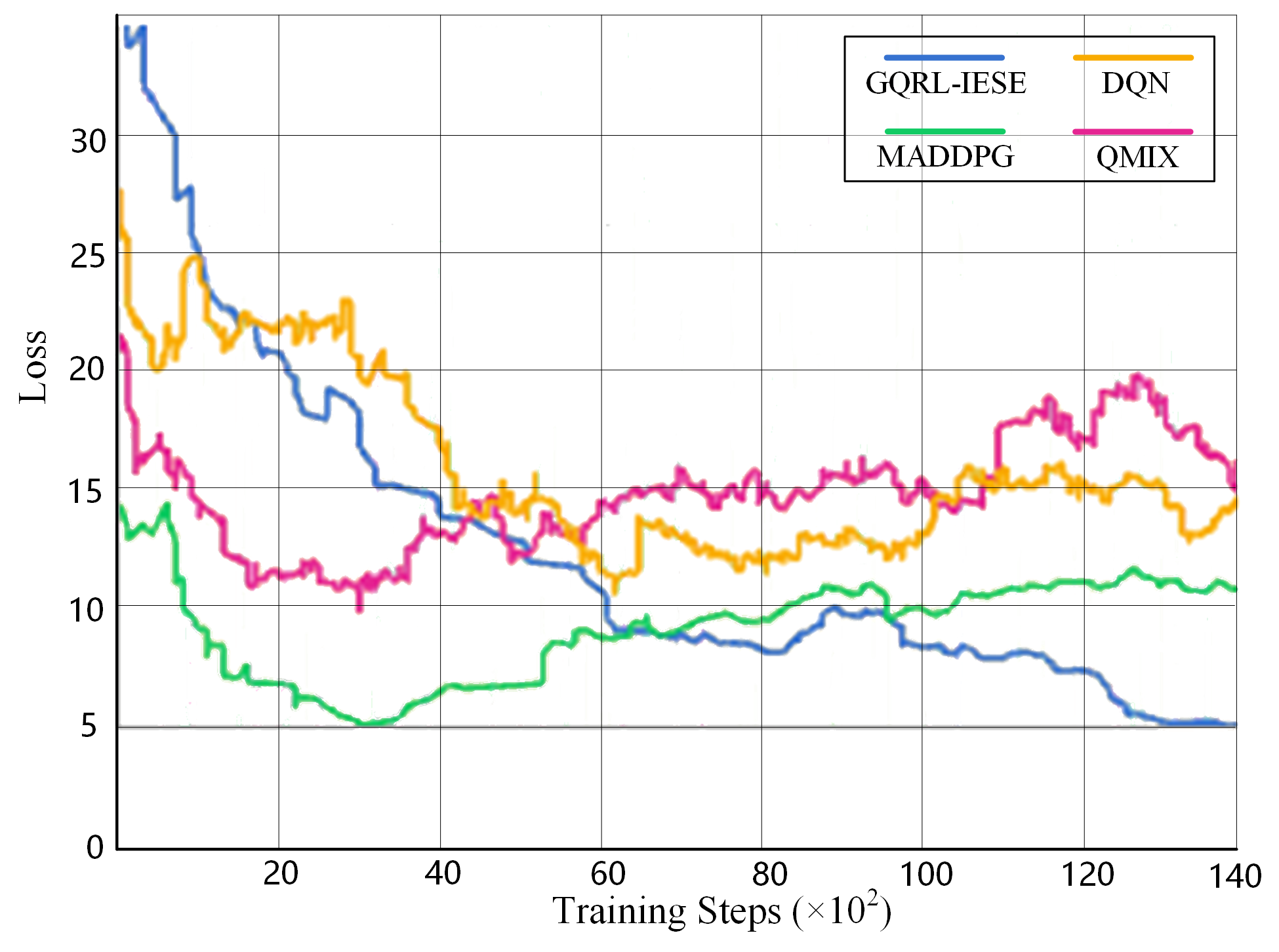}}
\caption{The loss curves of GQRL-IESE, QMIX, MADDPG and DQN}
\label{fig_4}
\end{figure}

Fig. \ref{fig_4} depicts the loss curves of different methods to compare their convergence. In Fig. \ref{fig_4}, the loss of GQRL-IESE uses the loss of the pursuing vehicles' decision network DQNs, and MADDPG uses the loss of \emph{Actor} network. It is evident that the proposed GQRL-IESE has the best convergence rate with the lowest loss value. MADDPG shows the second-best convergence rate and QMIX has the worst convergence performance. The best convergence performance of GQRL-IESE shows that it has a stronger ability to handle complex scenes and has excellent stability.

\begin{figure}[h]
\centering
\centerline{\includegraphics[width=0.4\textwidth,height=4.7cm]{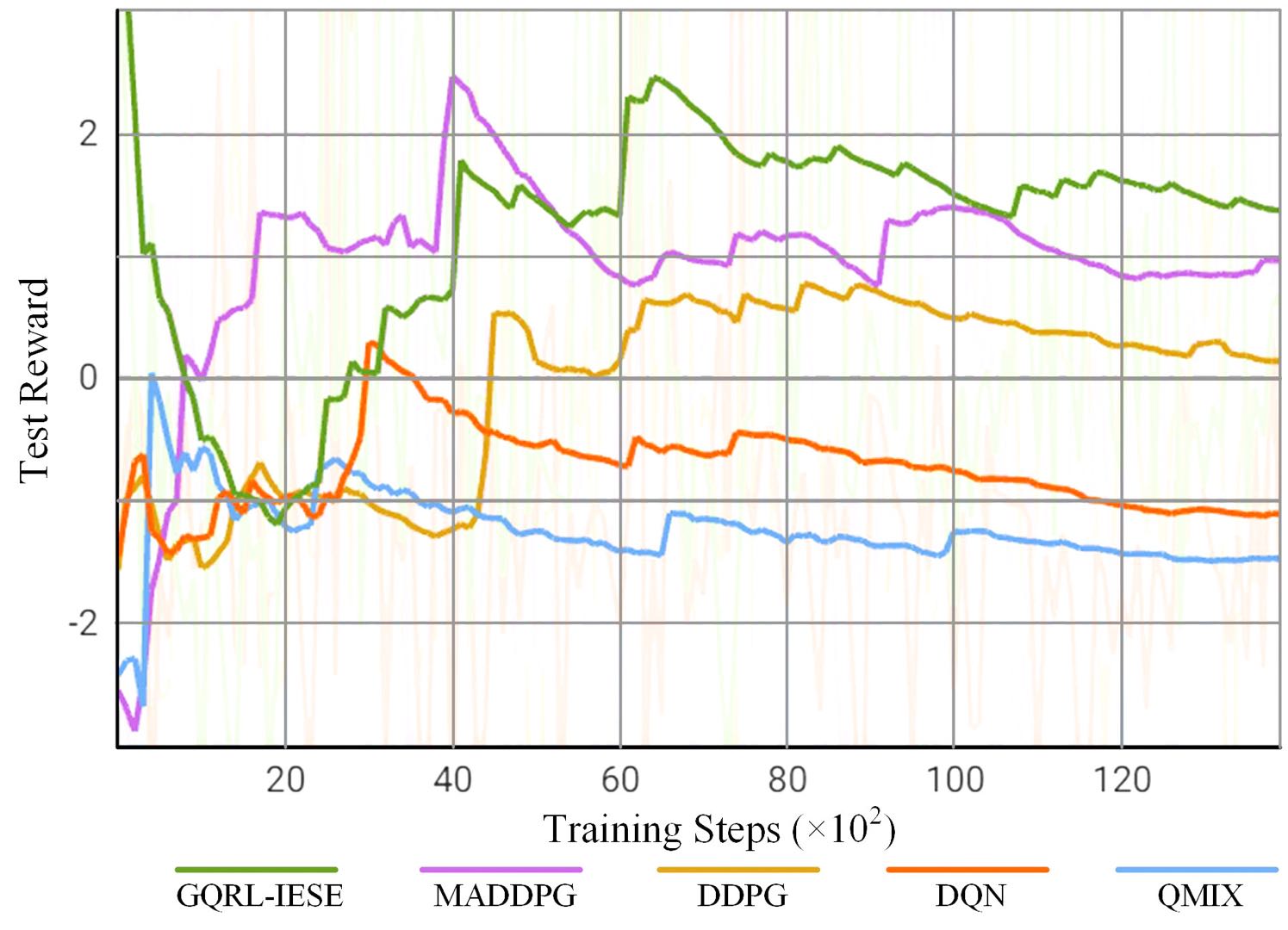}}
\caption{The test reward of GQRL-IESE, QMIX, DDPG, MADDPG and DQN}
\label{fig_5}
\end{figure}

As shown in Fig. \ref{fig_5},  it is intuitive that the GQRL-IESE shows apparent superiority over other methods. In particular, the proposed GQRL-IESE obtains the positive rewards, while the average rewards obtained by QMIX and DQN at each timestep are below $0$. 
This indicates that GQRL-IESE facilitates the pursuing vehicles to approach the evading vehicles more quickly, thereby improving the pursuing efficiency.
Although MADDPG performs better than GQRL-IESE initially, as the number of training timesteps increases, MADDPG becomes more volatile and the average reward at each timestep obtained gradually decreases, finally achieving second-best performance. This illustrates that the proposed GQRL-IESE achieve a more stable pursuing performance and superior pursuing efficiency.

Moreover, in order to more convincingly prove the superiority of the proposed GQRL-IESE scheme, we show the detailed data of three aspects of the total timesteps, the average reward at each timestep, and the total reward, which is shown in Table \ref{table2}.
The “Improvement with GQRL-IESE” in the fourth and sixth lines in Table \ref{table2} refer to the improvement of GQRL-IESE and other methods in total reward and average reward, respectively.
The total timestep of GQRL-IESE is 47.64$\%$ less than other methods on average, specifically, 61.54$\%$, 33.71$\%$, 18.98$\%$ and 56.47$\%$ less than that of QMIX, DDPG, MADDPG and DQN, respectively.
The total reward of GQRL-IESE is increased by 10.88$\%$ compared to MADDPG which has the second-best performance. And the average reward of GQRL-IESE at each timestep is 104.83$\%$ higher than all other methods on average.


\begin{figure}[htbp]
\centerline{\includegraphics[width=0.4\textwidth,height=4.7cm]{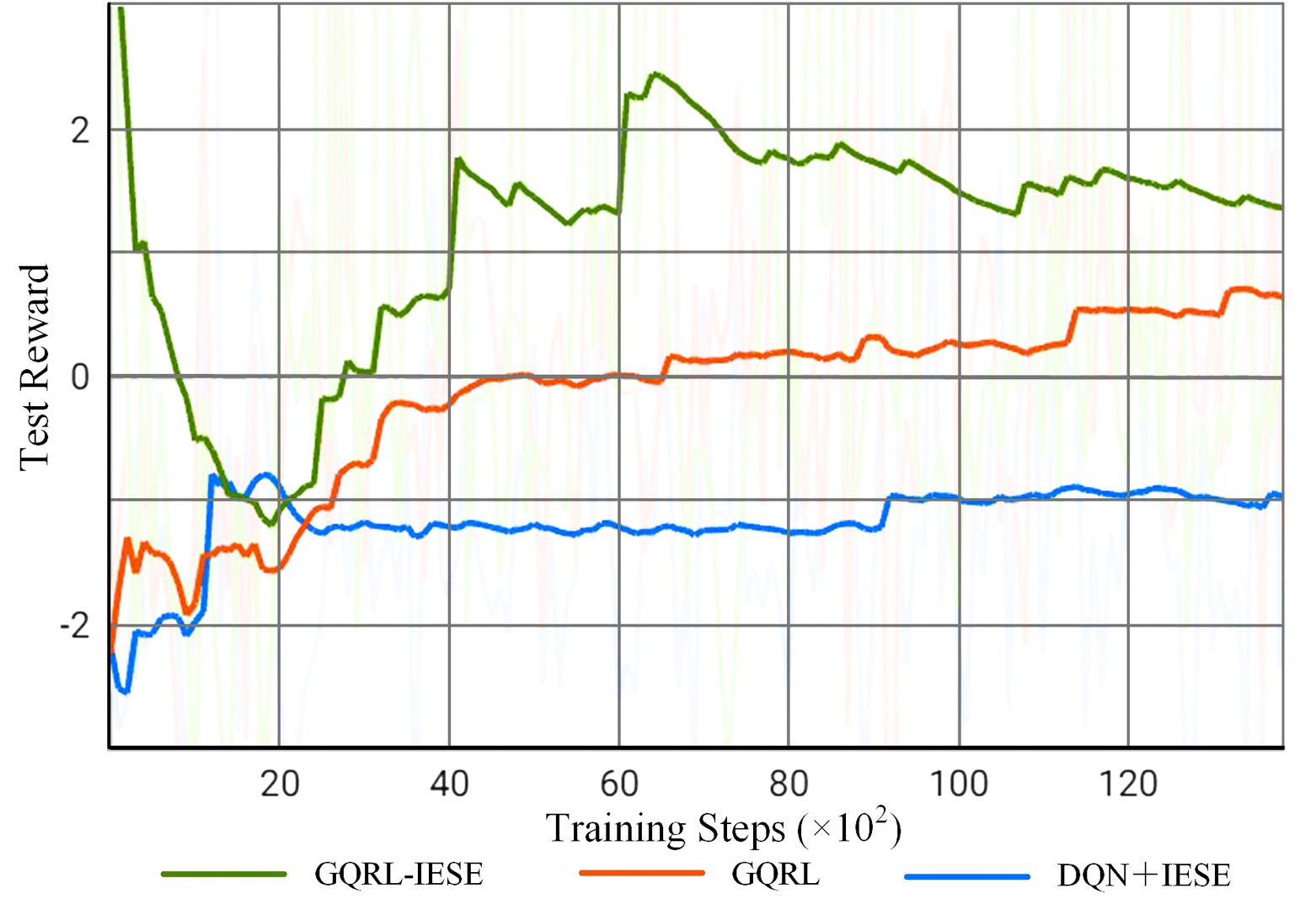}}
\caption{The test reward of GQRL-IESE, IESE+DQN and GQRL}
\label{fig_6}
\end{figure}

\subsection{Ablation Experiments}
Ablation experiments are performed to investigate the contributions of the IESE and cooperative graded-Q scheme in the proposed GQRL-IESE, respectively. GQRL represents the method that does not use IESE to process the state of the agents. IESE+DQN represents the method that does not use the coordinated Q optimizing network, which means that each agent only uses DQN to make decisions and execute directly. As shown in Table \ref{table2}, the total timestep and total reward of GQRL are increased by 68.00$\%$ and decreased by 54.20$\%$, respectively, compared to GQRL-IESE. This illustrates that IESE expedites the process of each pursuing vehicle to determine its own pursuing target, and helps the agent obtain richer input information, thereby improving the pursuing efficiency. Compared with GQRL-IESE, the total reward and average reward of IESE+DQN are reduced by 253.34$\%$ and 156.26$\%$, respectively, and the total timestep is increased by 172.57$\%$.
This demonstrates that the proposed cooperative graded-Q model effectively enables the pursuing vehicles to learn cooperative strategies from the global level, and improves the cooperation among the pursuing vehicles.

More intuitively, Fig. \ref{fig_6} depicts the performance of GQRL, IESE+DQN and GQRL-IESE. It is clear that GQRL-IESE achieves the best performance. Although IESE+DQN and GQRL have faster convergence rates, GQRL-IESE achieves more stable and effective pursuing performance.
The ablation experiments demonstrate that the proposed GQRL-IESE effectively improves the pursuing performance.

\section{Conclusion and Future Works}\label{VI}
This paper proposes a GQRL-IESE framework to solve the HCMVP problem under the complex urban traffic environment.
The IESE is proposed to encode the pursuing vehicles' state, which effectively removes redundant information received by the pursuing vehicles, and assists the pursuing vehicle in quickly determining its current pursuing target, thus promoting the DQN-based pursuing decision-making.
Moreover, a cooperative graded-Q scheme is proposed in GQRL-IESE to facilitate cooperation among the pursuing vehicles.
The coordinated Q optimizing network introduced into the cooperative graded-Q scheme, greatly facilitates the collaboration among pursuing vehicles and the collaborative optimization of decision-making. 
Extensive experimental results based on SUMO show that the proposed method is more competitive than existing methods. The total timestep of the proposed GQRL-IESE is less than other methods on average by 47.64$\%$.
The future work will focus on obtaining more intelligent evading vehicles to inversely facilitate cooperation among pursuing vehicles, and achieving more competitive performance in real-world traffic environments.


\bibliographystyle{unsrt}
\bibliography{msn_ref}

\end{document}